\documentclass[sigconf,screen]{acmart}

\AtBeginDocument{%
  \providecommand\BibTeX{{%
    \normalfont B\kern-0.5em{\scshape i\kern-0.25em b}\kern-0.8em\TeX}}}

\setcopyright{acmlicensed}
\copyrightyear{2024}
\acmYear{2024}
\acmDOI{XXXXXXX.XXXXXXX}

\acmConference[Preprint]{Preprint}{February 08,
  2024}{}
\acmISBN{XXX-X-XXXX-XXXX-X/XX/XX}

\usepackage{multirow}
\usepackage[ruled]{algorithm2e}
\usepackage{subcaption}

\begin{document}

\newcommand{\images}{\mathcal{I}}
\newcommand{\image}[1]{{I_{#1}}}
\newcommand{\curlybraces}[1]{ \{ #1 \} }
\newcommand{\crops}[1]{{X_{#1}}}  
\newcommand{\crop}[2]{{x^{#1}_{#2}}}  
\newcommand{\cropFeat}[2]{{z^{#1}_{#2}}}  
\newcommand{\groups}{\mathcal{B}}
\newcommand{\group}[1]{{B_{#1}}}
\newcommand{\testimages}{\mathcal{T}}
\newcommand{\testimage}[1]{{T_{#1}}}  
\newcommand{\bags}{\mathcal{B}}
\newcommand{\bag}[1]{{B_{#1}}}
\newcommand{\CEloss}{{\mathcal{L}_{\text{CE}}}}
\newcommand{\tripletloss}{{\mathcal{L}_{\text{triplet}}}}
\newcommand{\alignloss}{{\mathcal{L}_{\text{align}}}}

\title{Contrastive Multiple Instance Learning for Weakly Supervised Person ReID}

\author{Jacob Tyo}
\email{jtyo@cs.cmu.edu}
\affiliation{%
  \institution{DEVCOM Army Research Lab}
  \institution{Carnegie Mellon University}
  \streetaddress{5000 Forbes Avenue}
  \city{Pittsburgh}
  \state{Pennsylvania}
  \country{USA}
  \postcode{15239}
}

\author{Zachary C. Lipton}
\affiliation{%
  \institution{Carnegie Mellon University}
  \streetaddress{5000 Forbes Avenue}
  \city{Pittsburgh}
  \state{Pennsylvania}
  \country{USA}
  \postcode{15239}
}

\renewcommand{\shortauthors}{Tyo, et al.}

\begin{abstract}
  The acquisition of large-scale, 
  precisely labeled datasets for person re-identification (ReID) 
  poses a significant challenge. 
  Weakly supervised ReID has begun to address this issue, 
  although its performance lags behind fully supervised methods. 
  In response, 
  we introduce Contrastive Multiple Instance Learning (CMIL), 
  a novel framework tailored for more effective weakly supervised ReID. 
  CMIL distinguishes itself by requiring only a single model 
  and no pseudo labels, 
  while leveraging contrastive losses -- 
  a technique that has significantly enhanced 
  traditional ReID performance yet is absent in all prior MIL-based approaches. 
  Through extensive experiments and analysis across three datasets, 
  CMIL not only matches state-of-the-art performance 
  on the large-scale SYSU-30k dataset 
  with fewer assumptions 
  but also consistently outperforms all baselines 
  on the WL-market1501 and Weakly Labeled MUddy racer 
  re-iDentification dataset (WL-MUDD) datasets. 
  We introduce and release the WL-MUDD dataset, 
  an extension of the MUDD dataset
  featuring naturally occurring weak labels 
  from the real-world application at \url{PerformancePhoto.co}.
  All our code and data are accessible at \url{https://drive.google.com/file/d/1rjMbWB6m-apHF3Wg_cfqc8QqKgQ21AsT/view?usp=drive_link}. 
\end{abstract}

\ccsdesc[500]{Computing methodologies~Semi-supervised learning settings}
\ccsdesc[500]{Computing methodologies~Visual content-based indexing and retrieval}

\keywords{Multiple Instance Learning, Weak Supervision, Contrastive Learning, Person Re-Identification, Machine Learning}



\maketitle

\section{Introduction}

Accurate data labeling is a critical part of any machine-learning system, 
but is often prohibitively expensive, 
especially for person re-identification (ReID). 
In most classification problems, 
the classes are easily human-recognizable, 
allowing annotators to quickly recognize and label the class of a data point. 
In ReID however, 
the data points can consist of millions of individuals, 
none of which are known to the annotators. 
In this case, 
generating accurate labels is extremely difficult
and time-consuming. 
An alternative approach is to use 
weakly supervised learning (WSL) methods
that can effectively leverage lower-quality data labeling,  
which is often available in larger amounts 
at meager cost~\citep{liu2023weakly, wang2021learning, meng2019weakly, zhao2021weakly, wang2020weakly}. 

WSL has achieved impressive results on benchmark datasets, 
but performance still lags that of the standard, fully-supervised, setting.
Given that the reid task of identifying images of the same person is inherently contrastive with respect to identities, it seems possible that we could leverage techniques from contrastive learning to improve WSL further. 
Contrastive learning is a specific subset of supervised learning where models are optimized on pairs (or triplets) of inputs to determine if the inputs originate from the same class or not. 
This slight reframing has major benefits both in terms of model 
performance and generalization~\citep{garg2023complementary, hermans2017defense}, 
and in terms of computational efficiency in downstream applications. 
Specifically for ReID, 
common applications include facial recognition, 
person search, and image retrieval. 
In each of these settings, 
the number of downstream classes (identities) is typically unknown, 
a setting where contrastive models excel. 
However, traditional algorithms in contrastive learning depend on accurate labels. 

Weak labels for ReID can be gathered in several ways. 
One example, as provided by \citet{guillaumin2010multiple}, 
is to gather images of people based on an online search. 
The resulting dataset is bags of images
that all contain the same person, 
but the images would also be extremely noisy, 
containing many other people in each photo. 
Another example of this type of weak labels
is to observe event photo purchases - 
someone purchasing photos of a racer after a marathon 
likely purchase photos that all contain a common single person. 
As part of this work, 
we introduce and release the Weakly Labeled MUddy racer 
re-iDentification dataset (WL-MUDD) dataset, 
which is a dataset labeled in this exact manner
from the motorcycle racing event photo website \url{PerformancePhoto.co}. 

The dominant methods for weak ReID 
rely on pseudo-labeling~\citep{liu2023weakly, zhao2021weakly, zheng2021weakly, wang2020weakly}, 
which is an iterative process of predicting new labels 
for the weakly labeled data in an attempt to build
better models. 
Other approaches include graph-based methods~\citep{wang2021learning, meng2021deep},
Multiple Instance Learning (MIL)~\citep{huang2017salient, wu2015deep, sudharshan2019multiple}, 
or transferring an unsupervised model (i.e. trained without labels).
The unsupervised methods have made significant progress recently, 
but still fall short of methods that can leverage labeling~\citep{wang2020weakly}. 
Within WSL, 
pseudo-labeling approaches typically outperform 
those of noisy learning and MIL. 
However, the existing MIL formulations 
restrict the use of contrastive methodologies. 

In this work, 
we introduce Contrastive Multiple Instance Learning
to enable contrastive learning among weakly labeled bags of images. 
Contrastive learning is typically interpreted as 
decreasing the distance between the representation 
of two images of the same identity (or class), 
and increasing the distance between the representation 
of two images of different identities. 
However, 
in weakly supervised learning, 
the labels are not that granular. 
Pseudo-labeling methods get around this
by trusting that the labels are granular enough, 
and then updates the labels as training progresses, 
but this is prone to errors. 
Especially in settings where the intra-identity variability is extremely high, 
and the inter-identity variability is low, 
which is the exact case for our WL-MUDD dataset. 
Instead of a label refinement approach, 
we focus on the MIL formulation, 
and to enable contrastive techniques, 
we formulate the contrastive learning problem as 
decreasing the distance between two \emph{bag} representations
that have the same label, 
and increase the distance between two \emph{bag} representations
with a different label. 

This shift in perspective, 
of optimizing for bag representations instead of representations of a single 
image within that bag, 
is not obvious, 
mainly because at test time, 
the goal is still to produce a high-performing ReID model: 
one that can take a single image and produce a high-quality embedding
for it. 
To this end, 
CMIL includes two processes to help in this regard.
The first is that 
each image in each bag is independently embedded
into a representation using a feature extraction network, 
resulting in a bag of image features. 
Then, 
the bag of image features is passed through an 
accumulation network
to generate a bag representation. 
Second, 
we experiment with an \emph{alignment loss}, 
to encourage our model to learn image and bag representations
that are similar. 

The feature extraction network is chosen to be a standard ReID model, 
specifically ResNet-50~\citep{he2016deep}. 
The accumulation network must be permutation invariant, 
and is therefore chosen to be a set transformer, 
although we do provide ablation studies with 
the simpler choices of the average, max, and sum operators. 
Surprisingly, 
we find that even without the alignment loss, 
optimizing for high-quality bag representations
implicitly leads to high-quality image representations. 

We evaluate CMIL against a state-of-the-art weakly supervised learning method~\citep{ye2021collaborative} and a prior MIL method~\citep{meng2019weakly} on the weakly-labeled Market1501 (WL-Market1501) dataset, and WL-MUDD datasets. 
The WL-Market1501 dataset is the widely used Market1501 dataset~\citep{zheng2015scalable} but with noise added to mimic the weakly labeled setting.
Then, 
we compare CMIL to the state-of-the-art 
on the large-scale SYSU-30k weakly labeled ReID dataset,
containing nearly 30 million images and over 30 thousand identities.
We find that on both WL-Market1501 and WL-MUDD, 
CMIL consistently achieved the best rank-1, 
rank-5, rank-10 accuracy, and mean accuracy precision. 
On the SYSU-30k dataset, 
CMIL matched the state-of-the-art while requiring fewer modeling assumptions. 
Lastly, 
our ablation studies reveal the surprising effectiveness of average pooling 
for image aggregation, 
along with the surprisingly different instance and bag representations
even of the best-performing models.

The contributions of this work are threefold: 
\begin{itemize}
    \item The introduction and release of WL-MUDD, 
    a real-world dataset of motorcycle racers with naturally weak labels from \url{PerformancePhoto.co}.
    \item We introduce CMIL, 
    a novel framework for re-identification from weakly labeled group images.
    \item Experimental evidence of the efficacy of CMIL and an analysis highlighting the surprising differences between image and bag representations. 
\end{itemize}

\begin{figure*}[t]
    \centering
    \includegraphics[width=\textwidth]{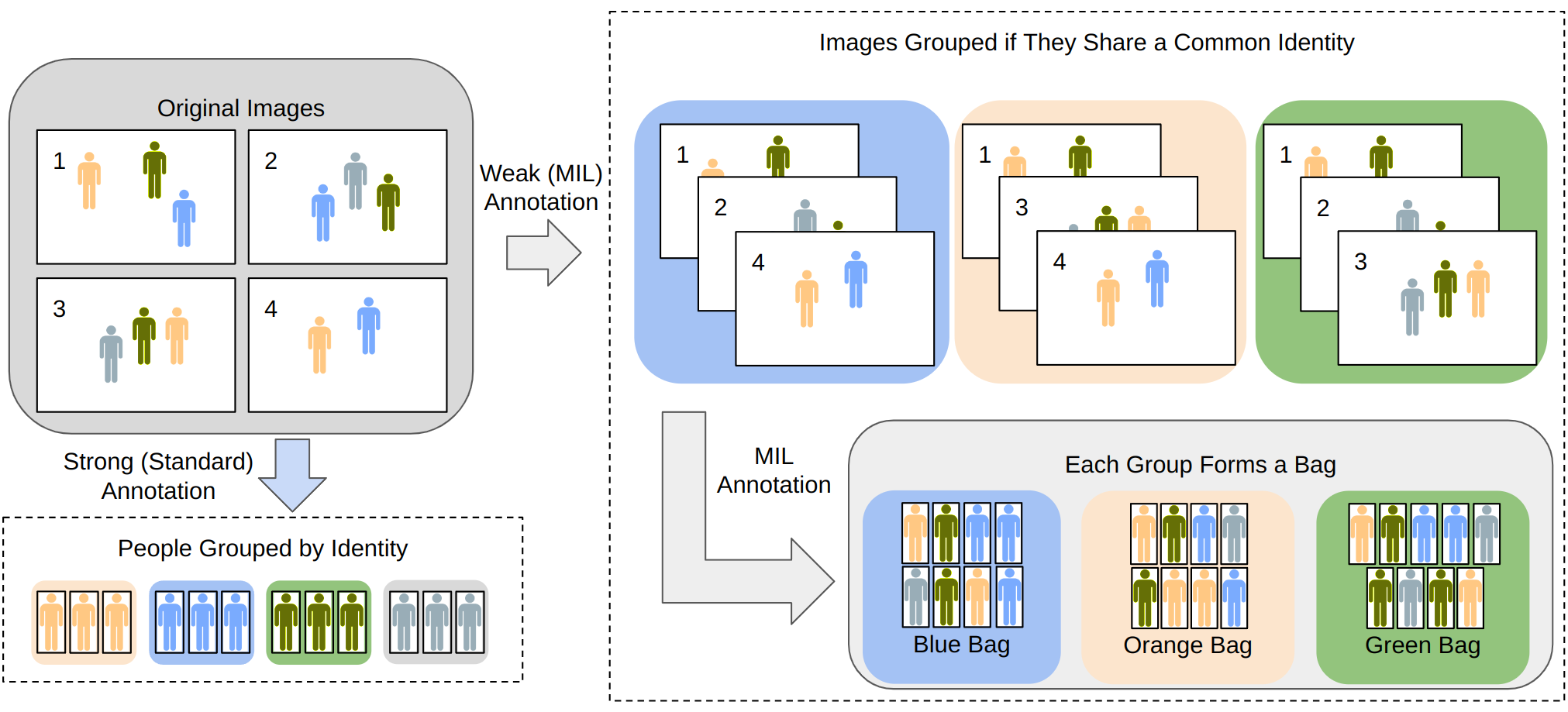}
    \caption{The annotation process for strong and weak ReID. The strong annotations group each crop into a bag based on their identity, whereas the weak annotation groups all images based on a shared identity, 
    and then all crops from the grouped images become a bag.}
    \label{fig:overview}
\end{figure*}

\section{Datasets and Problem Setup}

In this section, 
we formally introduce the weakly supervised ReID setting, 
as well as a new weakly supervised ReID dataset.
The dataset is available at \url{https://drive.google.com/file/d/1rjMbWB6m-apHF3Wg_cfqc8QqKgQ21AsT/view?usp=drive_link}.  

\subsection{Weakly Supervised Re-Identification}\label{sec:ws_rid}

The problem we address in this paper is re-identification 
from weakly labeled group images. 
We assume that we are given a dataset of images
$\images = \curlybraces{ \image{1}, \image{2}, \dots, \image{N} }$
where each image $\image{j}$
contains one or more people that we are interested in identifying. 
Let 
$\crops{j} = \curlybraces{\crop{j}{1}, \crop{j}{2}, \dots, \crop{j}{M_j}}$ 
denote the set of $M_j$ \emph{crops} containing each person
extracted from image $\image{j}$.
We refer to each specific person in an image $\image{j}$ as 
$\crop{j}{i} \in \crops{j}$. 

However, 
unlike conventional re-identification datasets, 
we only have weak labels for each image. 
These labels merely indicate the presence of a shared identity 
within each group, 
but not the specific identity of each individual instance within the group. 
This means that we have access to a set of \emph{bags}
$\bags = \curlybraces{\bag{1}, \bag{2}, \dots, \bag{K}}$
where each bag 
$\bag{k} = \curlybraces{\image{k_1}, \image{k_2}, \dots, \image{k_{|\group{k}|}}}$
contains images that share a common identity.
Importantly, 
the individual instances within each group are not labeled with their specific identities. 
Instead, the bag is labeled with only a single identity.
The key challenge in this setting is to learn a model 
that can effectively discriminate between different identities 
despite only having access to these weak bag-level labels. 
During inference time the bag-level labels are not the label of interest. 
Instead, we want a standard ReID model at inference time, meaning that we need to be able to predict the identity of a single person (i.e. crop) within an image, not of the bag. 

\subsection{WL-MUDD Dataset}
\label{sec:wl-mudd}

\url{PerformancePhoto.co} is an online marketplace for off-road racing 
photographers and fans. 
Powered by text spotting and ReID models to enable searchable racing photos, 
improvements to the ReID models suffer from the high costs 
of ReID dataset labeling. 
However, 
there is a proxy that gives natural weak labels: user purchases. 
When a user purchases photos from a single event, 
they are likely purchasing photos of a single individual. 
However, 
it is also likely that there is more than one individual in each photo purchased. 
Following notation from Section~\ref{sec:ws_rid}, the set of photos purchased by a single user can be regarded as a \textit{bag}
that can be weakly labeled with a unique identity. 

The MUDD~\citep{tyo2023mudd} dataset was curated from \url{PerformancePhoto.co} and manually labeled in the traditional, 
fully supervised, ReID setting. 
We adapt the MUDD dataset to the weakly supervised setting by re-labeling the data points at the bag level and adding all, previously unlabeled, crops to each bag according to their existence in the original images.
Figure~\ref{fig:overview} shows this labeling process 
and compares it to the standard (strong) annotation procedure. 
Instead of relying on the user purchases heuristic, 
we were able to build out the WL-MUDD dataset 
by taking all of the strong labels from the MUDD dataset, 
and then linking them back to the photos they originated from. 
Then, 
we take all the other people in the original photo, 
and add them to the dataset under the same label, 
forming a bag.
This is repeated for every person in the original MUDD dataset,
resulting in a weakly labeled dataset over twice as large. 
We refer to this dataset as the Weakly Labeled MUddy racer 
re-iDentification dataset (WL-MUDD).

\begin{figure*}[t]
    \centering
    \begin{subfigure}[b]{0.24\textwidth}
        \centering
        \includegraphics[width=\textwidth]{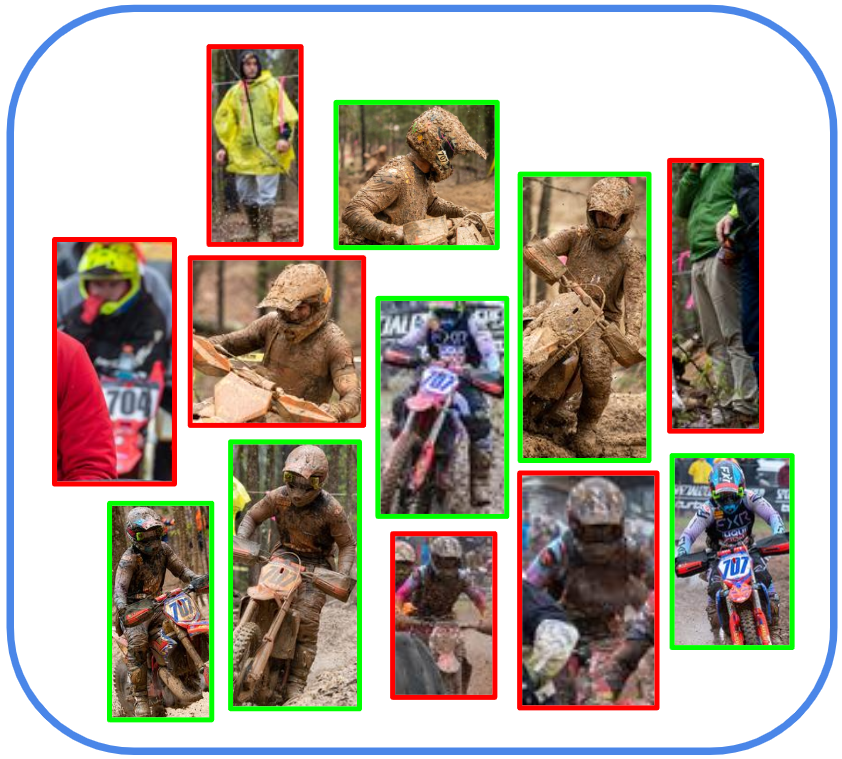}
    \end{subfigure}
    \hfill
    \begin{subfigure}[b]{0.24\textwidth}
        \centering
        \includegraphics[width=\textwidth]{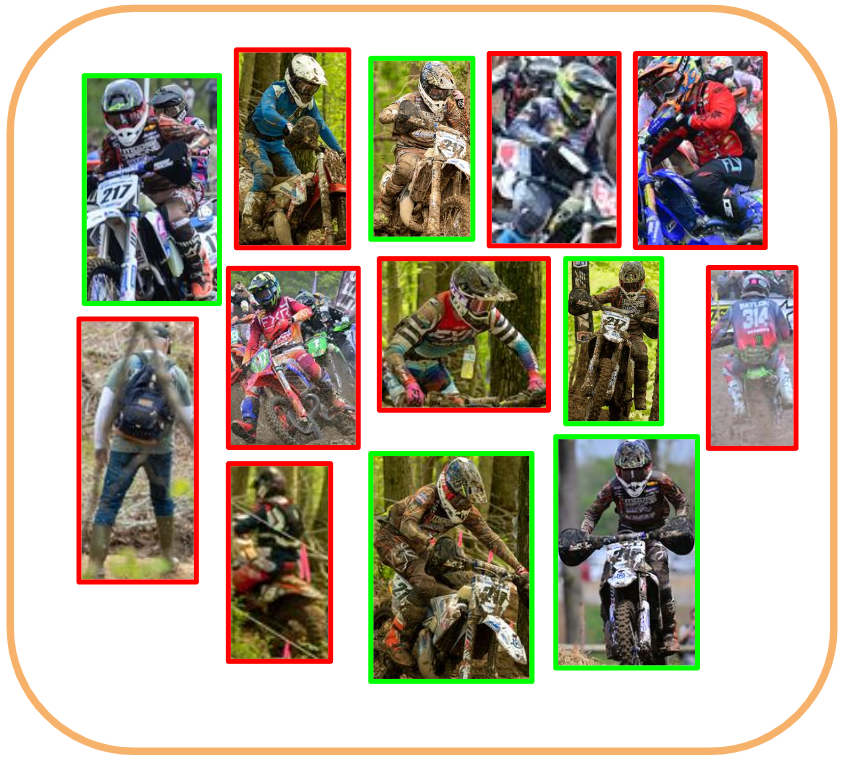}
    \end{subfigure}
    \hfill 
    \begin{subfigure}[b]{0.24\textwidth}
        \centering
        \includegraphics[width=\textwidth]{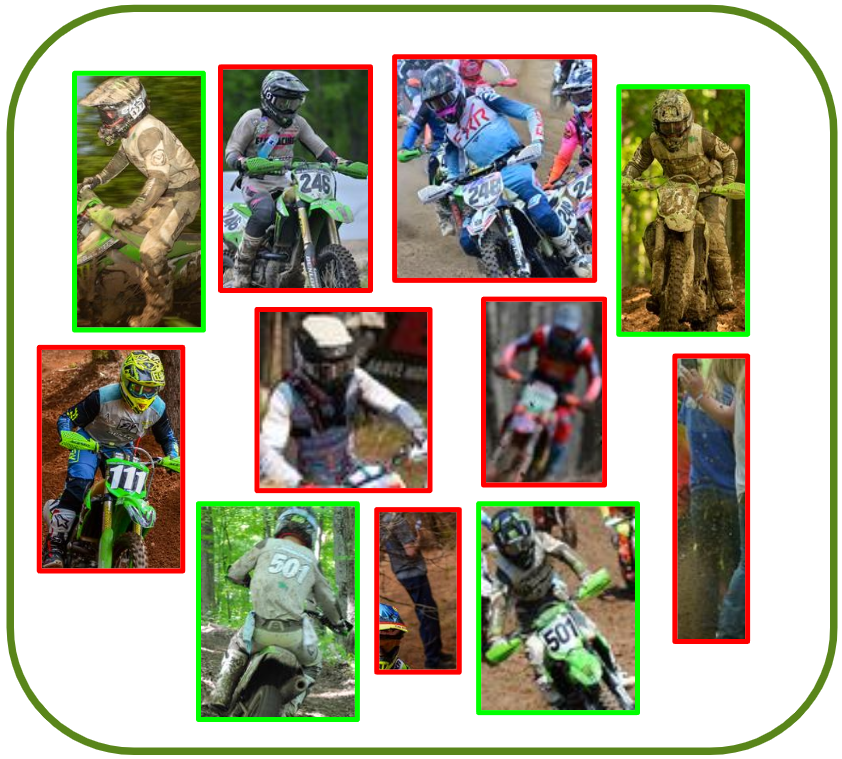}
    \end{subfigure}
    \hfill
    \begin{subfigure}[b]{0.24\textwidth}
        \centering
        \includegraphics[width=\textwidth]{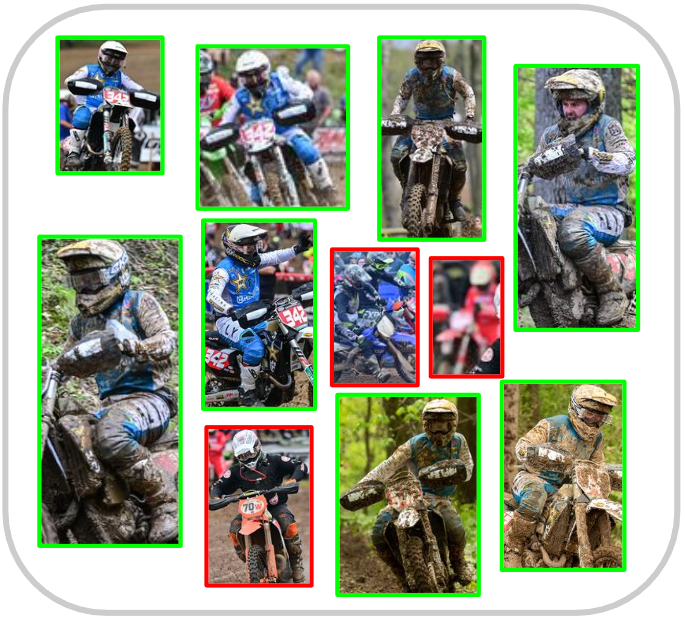}
    \end{subfigure}
    \caption{Four example subsets from four different bags of the WL-MUDD dataset. 
    Each image within a bag is outlined in green if it is the same identity as the bag, 
    and red if it is not. Each bag can have very different ratios of correct to incorrect identities of the underlying images.}
    \label{fig:wl-mudd-examples}
\end{figure*}

The average bag in WL-MUDD has 75 crops of people in it, 
with 32\% of them being the identity of the label attributed to that bag.
This corresponds to an average noise level of 68\%. 
The bags can be as small as 5 crops, 
or as large as 300, 
and the noise level of each bag varies between 50\% and 85\%. 
Figure~\ref{fig:wl-mudd-examples} gives examples of bags in the dataset, 
highlighting the extremely high inter-class variation. 
The crops highlighted in green are representative of the bag label, 
whereas the red highlighted crops are not.

\section{Contrastive Multiple Instance Learning}

\begin{figure*}[t]
    \centering
    \includegraphics[width=\textwidth]{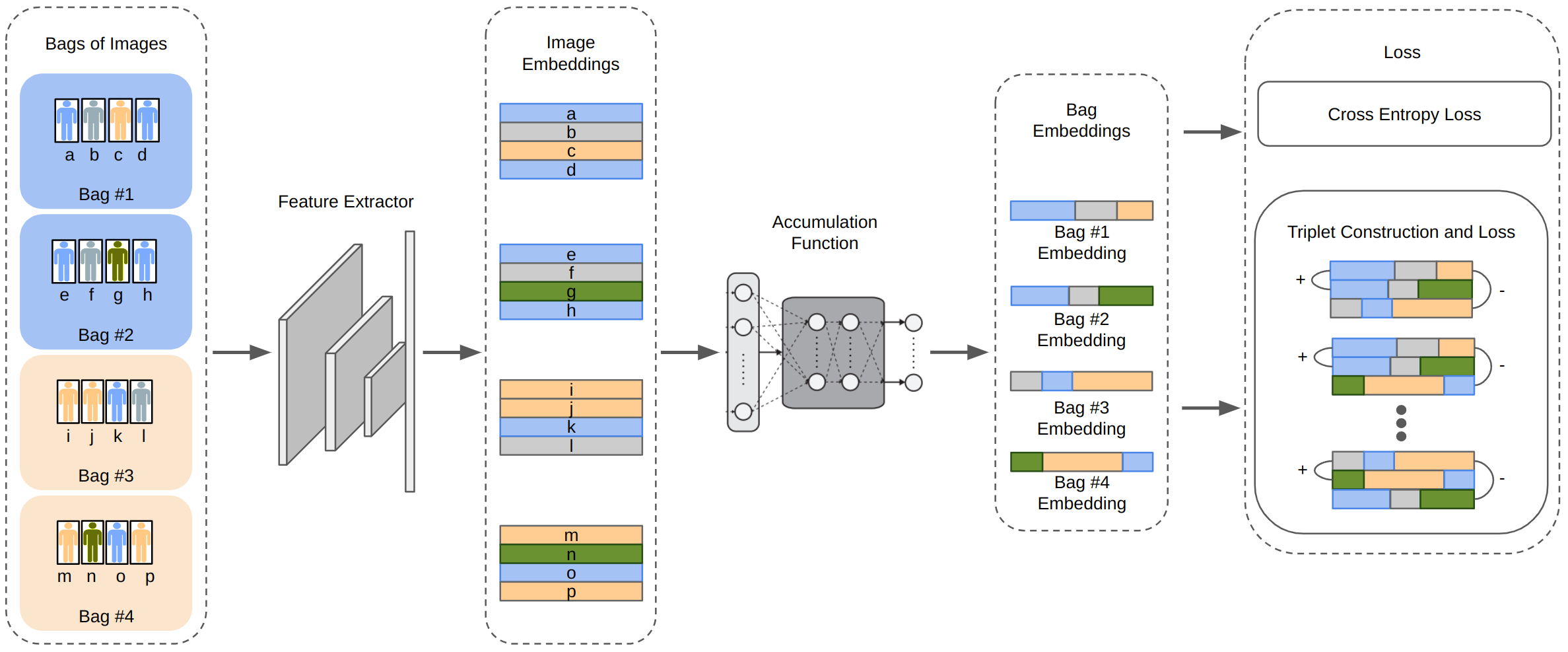}
    \caption{The CMIL framework. For each image in a batch of bags, a feature extraction network is used to get an embedding for each image. Then for each bag, the corresponding image embeddings are combined into a single bag embedding via an accumulation function. Finally, the bag embeddings are used to calculate the cross entropy loss (or identity loss), as well as the triplet loss based on all valid triplets from the batch.}
    \label{fig:CMIL-model}
\end{figure*}

We cast the weakly supervised object re-identification problem
as one of multiple-instance learning
and present the contrastive multiple-instance learning (CMIL) method. 
A standard multiple-instance learning problem handles bag-level labeling 
by getting a feature representation for all crops in a bag,
applying an accumulation function (typically max, average, etc.) to get a single
bag representation from all of the crop representations, 
and then applying a classifier to the bag representation to determine a classification. 
Instead of a bag classifier (or alongside), 
we compare bag representations via a contrastive loss. 
This allows us to train end-to-end in a contrastive fashion. 
The CMIL framework is shown in Figure~\ref{fig:CMIL-model}.

This is a divergence from standard contrastive learning.
At test time we compare representations of crops, 
and therefore the goal of training is to optimize the crop representations accordingly. 
However, 
in this formulation, 
we are directly optimizing the bag representations, 
and only indirectly optimizing the crop representations. 
Specifically,
all crops from a single bag $k$ are encoded into crop representations 
by a model $f$ parameterized by $\theta$. 
\begin{align}
    \cropFeat{i}{j} = f_\theta (\crop{i}{j}), \ \forall j \in M_i, i \in N_k,
\end{align}
where $M_i$ is the number of crops in image $i$ and $N_k$ is the set of images in bag $k$. 
It is critical that this model takes a specific crop as input, 
and returns the corresponding representation for that input
because during testing, 
this is the only aspect of the model that will be utilized. 
Then given all crop representations for a bag $k$, 
they must be accumulated into a single bag-representation 
using a model $g$ parameterized by $\phi$. 
\begin{align}
    r_k = g_\phi ( \cropFeat{1}{1}, \dots, \cropFeat{N_k}{M_i} ).
\end{align}
This accumulation function should be permutation invariant to the input, 
as there is no way to control the ordering of the instances meaningfully. 

The final component of this architecture is a distance or similarity function $d$. 
To apply contrastive learning,
we must be able to measure the distances between pairs/triplets/quadruplets of bags. 
Any proper distance metric, 
such as the Euclidean or cosine distance, 
can be used. 
This distance can then be used to return a ranking, 
thresholded to provide a classification, etc.
We focus on the setting where we are given a triplet of bags
(or by the time it reaches the distance metric, bag representations). 
Given a bag $a$ and $b$, the distance between their representations is represented by:
\begin{align}
    \hat y = d(r_a, r_b). 
\end{align}

Note that in this methodology,
we depend on bags of data during each iteration. 
Each iteration must have a sufficient number of bags,
as well as a sufficient number of crops from each bag.
Therefore, 
the number of crops in each bag is intimately tied 
to both the batch size and the underlying assumptions 
about the nature (i.e. noise level) of the bags in the data. In most cases, 
the number of crops in a bag is large, 
and therefore we must sample mini-bags 
(e.g. a subset of a bag)
to construct a batch. 
If the bag sizes are too small, 
then it is likely that there will not be enough of the true
underlying identity in each bag to learn effectively. 
On the contrary, if the bags are too large, 
then it is likely that the number of bags in each batch 
is not sufficient for training. 
An implicit assumption of this framework
is that in expectation, 
the most common identity in 
each bag is the identity representative of the bag label. 
The noise level can still be high without violating this assumption, 
because most non-representative crops in a bag 
are of different identities altogether.
To ease notation, 
we will refer to bags and mini-bags interchangeably. 
In general, 
we mean mini-bags in algorithmic contexts, 
and bags in dataset contexts. 

During inference,
we follow the standard object re-identification procedure. 
Given a query set, gallery set, and an optional distractor set, 
we search for a specific object in the gallery based on a query image. 
All crops are embedded using our instance feature extractor, 
and then the distance metric used during training is used to return a ranking 
over the gallery and distractors for each query image. 
Based on this ranking, 
we track the rank-k accuracy for $k \in \curlybraces{1, 5, 10}$, 
as well as the mean average precision (mAP). 

\begin{algorithm}
\caption{Contrastive Multiple Instance Learning (CMIL)}\label{alg:cmil}
\textbf{Input:} Set of bags $\mathcal{B} = \{B_1, B_2, ..., B_N\}$, where each bag $B_i$ contains crops $\{x_1^i, x_2^i, ..., x_{M_i}^i\}$

\textbf{Output:} Trained model parameters $\theta$ for crop feature extraction

Initialize model parameters $\theta$, $\phi$, and $\psi$ randomly

\While{not converged}{
    \For{$\mathcal{B}_{batch} \subset \mathcal{B}$}{  
        \For{$B_i \in \mathcal{B}_{batch}$}{
            \For{$x_j^i \in B_i$}{
                $z_j^i = f_\theta(x_j^i)$ \# Extract features from crops
            }
            $r_i = g_\phi(\{z_1^i, ..., z_{M_i}^i)\})$ \# Aggregate crop features into bag representation
        }
        $\mathcal{L}_{triplet}(r_{1, \dots, |B_{batch}|})$ \# Triplet Loss
        
        $\mathcal{L}_{CE}(h_\psi(r_{1, \dots, |B_{batch}|})$ \# CE Loss
        
        $\alignloss(r_{1, \dots, |B_{batch}|}, z^{1, \dots, |B_{batch}|}_{1,\dots,|M_i|})$ \# Align Loss

        $\mathcal{L} = \alpha \tripletloss + \beta \CEloss + \gamma \alignloss$ \# Aggregated Loss
        
        Update model parameters $\theta$, $\phi$, $\psi$ to minimize $\mathcal{L}$
    }
}
\end{algorithm}

\subsection{Loss Function}

\begin{table}[h]
    \caption{Dataset Summary statistics for each dataset used in the experiments.}
    \label{tab:dataset-summaries}
    \begin{tabular}{@{}cccc@{}}
        \toprule
        Dataset      & Market-1501 & SYSU-30k       & weak MUDD \\ \midrule
        \# identities & 1,501       & 30,508         &  150     \\
        Scene        & Outdoor     & Indoor,Outdoor & Outdoor   \\
        Annotation   & Strong      & Weak           & Weak      \\
        Cameras      & 6           & Countless      & Countless \\
        Images       & 32,668      & 29,606,918     & 9,069     \\ \bottomrule
    \end{tabular}
\end{table}

\begin{table*}[t]
\centering
\caption{The sweep configuration for hyperparameter optimization, along with the final CMIL hyperparameters for each dataset. $U_{int}(x, y)$ represents an integer uniform distribution from $x$ to $y$, $U_{log}(x, y)$ represents a log uniform, and $U(x, y)$ represents a standard uniform distribution on all real numbers from $x$ to $y$.}
\label{tab:sweep-config}
\begin{tabular}{@{}ccccc@{}}
\toprule
\multirow{2}{*}{Parameter} & \multirow{2}{*}{Search Range} & \multicolumn{3}{c}{Final Values} \\
                      &                  & Market-1501 & SYSU-30k & Weak MUDD \\ \midrule
bag size        & $U_{int}(5, 10)$       & 6            & 5        & 9         \\
batch size      & $U_{int}(5, 10)$       & 10           & 10       & 5         \\
distance metric & [euclidean, cosine]    & cosine       & cosine   & cosine    \\
fixbase epoch   & $U_{int}(0, 10)$       & 7            & 10       & 8         \\
learning rate   & $U_{log}(1e-05, 0.01)$ & 2.1153e-4    & 2.828e-3 & 4.044e-4  \\
margin          & $U(0.1, 1)$            & 0.9992       & 0.8592   & 0.7731    \\
feature norm    & [false, true]          & False        & False    & False     \\
gamma           & [0, 0.01, 0.1]         & 0            & 0        & 0         \\
alpha           & $U(0, 1)$              & 0.5638       & 0.8083   & 0.3882    \\
beta            & $U(0, 1)$              & 0.3872       & 0.9242   & 0.7339    \\ \bottomrule
\end{tabular}
\end{table*}

\begin{table}[h]
\centering
\caption{Results on the WL-Market1501 dataset at varying levels of noise. The noise level represents the percentage of the dataset with incorrect labels. This dataset was synthetically constructed by duplicating images in the training set and assigning them to random bags -- 75\% noise would correspond to duplicating each image three times, therefore only 1 in 4 images would be correctly labeled.}
\label{tab:synth-results}
\begin{tabular}{@{}cccccc@{}}
\toprule
           Noise    & Method      & R1 & R5 & R10 & mAP \\ \midrule
\multirow{3}{*}{50\%} & CORE        & 80.9\% & 92.2\% & 95.0\% & 48.6\% \\
                    & MIML        & 71.8\% & 87.0\% & 91.5\% & 46.3\% \\
                    & CMIL (Ours) & 80.7\% & 91.9\% & 94.4\% & 56.8\% \\ \bottomrule
\multirow{3}{*}{66\%} & CORE        & 68.1\% & 83.6\% & 88.2\% & 38.6\% \\
                    & MIML        & 62.7\% & 82.2\% & 87.6\% & 38.8\% \\
                    & CMIL (Ours) & 76.4\% & 89.6\% & 93.0\% & 54.4\% \\ \bottomrule
\multirow{3}{*}{75\%} & CORE        & 56.1\% & 74.4\% & 80.8\% & 27.9\% \\
                    & MIML        & 50.4\% & 71.6\% & 79.31\% & 26.6\% \\
                    & CMIL (Ours) & 70.0\% & 86.4\% & 90.9\% & 48.8\% \\ \bottomrule
\multirow{3}{*}{80\%} & CORE        & 47.5\% & 66.0\% & 73.2\% & 17.4\% \\
                    & MIML        & 54.0\% & 60.8\% & 71.1\% & 19.0\% \\
                    & CMIL (Ours) & 64.9\% & 82.8\% & 88.0\% & 43.9\% \\ \bottomrule
\end{tabular}
\end{table}

CMIL leverages both the identity and triplet losses. 
The identity loss is the cross entropy loss when each class represents a person identity
\begin{align}
    \CEloss = - \sum_{c=1}^{C} y_{c} \log (p_c),
\end{align} 
where $y_c$ is a binary indicator (0 or 1) indicating the label of a sample for class $c$, 
and $p_c$ is the predicted probability of that class for the same sample, 
calculated by applying a fully connected layer 
($h$ parameterized by $\psi$)
and a softmax to the bag representation: 
\begin{align}
    p_i = \text{softmax}\Big( h_\psi ( r_i ) \Big). 
\end{align}
The triplet loss is: 
\begin{align}
    \tripletloss = \max \Big( d(r_a, r_p) - d(r_a, r_n) + m_\text{triplet}, 0 \Big) 
\end{align}
where $r_a$ is a bag representation for an anchor sample, 
$r_p$ is a bag representation for a positive sample 
(i.e. a bag with the same label as the anchor sample), 
and $r_n$ is a bag representation for a negative sample 
(i.e. a bag with a different label than the anchor sample). 

Again, 
this is explicitly optimizing bag representations
and only implicitly optimizing crop representations. 
In an attempt to address this, 
we experiment with an \emph{alignment loss}. 
The intuition is that the most shared identity in a bag
is the identity of interest. 
So an ideal accumulation function 
is one that can accurately pick out the representative crops, 
and then create a bag representation very similar to one or all of them. 
Therefore, 
we create the alignment loss to encourage the bag representation 
for a bag $a$ with crops $\curlybraces{\crop{a}{1}, \crop{a}{2}, \dots, \crop{a}{N_a}}$
to be close to any one of the crop representations:
\begin{align}
    \alignloss = \max \Big(0, \min \curlybraces{d(r_a, \cropFeat{1}{a}), d(r_a, \cropFeat{2}{a}), \dots, d(r_a, \cropFeat{N_i}{a})} - m_\text{align}),
\end{align}
where $m_\text{align}$ is a margin hyperparameter. 

The total loss function is a weighted combination of the 
identity, triplet, and alignment losses. 
The weighting for each loss (i.e. $\alpha$, $\beta$, and $\gamma$) 
is selected during our hyperparameter search. 
\begin{align}
    \mathcal{L} = \alpha \tripletloss + \beta \CEloss + \gamma \alignloss
\end{align}
Note that the triplet loss can be substituted with any contrastive loss. 
Algorithm~\ref{alg:cmil} provides an overview of CMIL in pseudocode.

\section{Experiments}

We evaluate our methodology on three datasets:
\begin{itemize}
    \item WL-Market-1501: 
    The widely used Market-1501 person 
    ReID dataset~\citep{zheng2015scalable}, 
    but with synthetically weak labels. 
    The synthetic labels are generated by duplicating
    images from the training set some number of times, 
    and assigning them to random bags. 
    \item WL-MUDD: Our real-world dataset introduced in Section~\ref{sec:wl-mudd}
    \item SYSU30k: A large-scale weakly supervised person ReID dataset 
    with over 29 million images gathered from TV program videos. 
    The videos are randomly broken into clips, 
    and then each clip is manually annotated with an identity, 
    but all detected people are noisily assigned that identity, 
    forming bag-level labels.
\end{itemize}
The dataset statistics
can be seen in Table~\ref{tab:dataset-summaries}. 
While the training set of each of these datasets is weakly labeled, 
the test sets are accurately labeled 
for normal person ReID evaluation~\citep{ye2021deep}.
We track the mean average precision (mAP)
and the Rank-k accuracy for $k \in \{ 1, 5, 10 \}$.

\subsection{Implementation Details and Hyperparameter Tuning}

\subsection{Baseline Methods}

\citet{ye2021collaborative} introduce online CO-REfining (CORE), 
a framework for online co-refining of ReID models.
CORE uses learning rate schedules to 
optimize two models collaboratively, 
while also iteratively refining the noisy labels in a dataset. 
CORE is a state-of-the-art method for learning ReID models 
among noisy labels and weak supervision.

\citet{meng2019weakly} introduce Cross View 
Multi-Instance Multi-Label Learning (CV-MIML). 
Being based on MIL, 
this method falls most closely related to ours. 
Although originally developed for the setting 
where a target person is known to appear within an untrimmed video
but no further information is available, 
this weakly supervised setting is equivalent to ours, 
although perhaps simpler due to correlations within a single video frame. 
Importantly, 
this method only performs bag classification during training, 
taking advantage only of the identity loss. 
Instead, CMIL optimizes bag representations explicitly.

We implement the CMIL framework using PyTorch. 
For a fair comparison, 
all methods utilize ResNet-50, 
pretrained on Imagenet~\citep{deng2009imagenet},
as the feature extractor $f_\theta$. 
Our method also requires a reduction function $g$, 
and in this case, we use a 2-layer set transformer~\citep{lee2019set}. 
Section~\ref{sec:ablation} includes ablations 
where we experiment with simpler reduction functions, namely the average,
max, and sum operators.
Importantly, 
we also implement a \emph{bag} sampling function. 
We expect two conditions to be met for every mini-batch: 
\begin{enumerate}
    \item Each batch will consist of $b$ sub-bags, 
    where a sub-bag is a subset of a bag. 
    If a bag is smaller than $b$, then the bag is oversampled.
    \item Each bag label present in the mini-batch will have two 
    or more bags in the mini-batch to ensure that valid triplets 
    can always be constructed.
\end{enumerate}

For hyperparameter selection, 
we run a Bayes hyperparameter search 
with early stopping (if model validation accuracy has not improved in 5 epochs, 
terminate the run) and 
hyperband (with an eta value of 2 and a minimum iteration count of 3)
for early termination of less promising runs~\citep{li2018hyperband}. 
Table~\ref{tab:sweep-config} describes the hyperparameter search ranges. 
The search aims to maximize the rank-1 accuracy 
on the validation set over 50 epochs.
For each dataset, 
250 models with hyperparameters sampled from the listed distributions
were trained and evaluated, 
and the best-performing hyperparameters are also shown in Table~\ref{tab:sweep-config}. 
Finally, using the best-performing hyperparameters, 
a final training run was done using the combined training and validation set, 
evaluated on the test set, and reported in our results.

\section{Results and Discussion}

Table~\ref{tab:synth-results} summarizes the rank-1 (R1), 
rank-5 (R5), rank-10 (R10) accuracy 
and mean average precision (mAP) of the different methods 
on the Market-1501 dataset with varying levels of synthetic label noise.
At the 50\% noise level, our CMIL method achieves an R1 accuracy of 80.7\%, 
nearly matching the performance of CORE and outperforming MIML by 8.9\%. 
As the noise level begins to increase, 
CMIL dominates the other methods by a growing margin. 
At 80\% label noise, 
the hardest setting, 
CMIL obtains a R1 of 64.9\%, 
which signifies a 10.9\% boost over the best baseline. 
The consistent gaps between CMIL and other approaches 
illustrate that our method can effectively 
learn useful representations among even extremely noisy bags.

Table \ref{tab:weak-mudd-results} 
summarizes the performance of different methods 
when trained on the real-world Weak MUDD dataset. 
With noisy group annotations, 
our CMIL framework obtains 73.2\% rank-1 accuracy. 
This significantly outperforms baseline methods, 
including CORE and MIML, by 2.5\% and 6\% respectively.

\begin{table}[h]
\centering
\caption{Results on the WL-MUDD dataset.}
\label{tab:weak-mudd-results}
\begin{tabular}{@{}ccccc@{}}
    \toprule
    Method      & R1 & R5 & R10 & mAP \\ \midrule
    CORE        & 67.2\% & 83.3\% & 92.3\% & 71.6\% \\
    MIML        & 70.7\% & 87.7\% & 95.2\% & 74.6\% \\
    CMIL (Ours) & 73.2\% & 90.0\% & 96.8\% & 75.1\% \\ \bottomrule
\end{tabular}
\end{table}

\begin{table}[h]
\centering
\caption{Results on the SYSU30k dataset.}
\label{tab:sysu30k}
\begin{tabular}{@{}ccc@{}}
\toprule
Supervision & Method                           & R1     \\ \midrule
\multirow{4}{*}{Transfer Learning} & DARIR~\citep{wang2016dari}         & 11.2\% \\
            & DF~\citep{ding2015deep}          & 10.3\% \\
            & Local CNN~\citep{yang2018local}  & 23.0\% \\
            & MGN~\citep{wang2018learning}     & 23.6\% \\ \midrule
\multirow{4}{*}{Self-Supervised}  & SimCLR~\citep{chen2020simple}      & 10.9\% \\
            & MoCo v2~\citep{chen2020improved} & 11.6\% \\
            & BYOL~\citep{grill2020bootstrap}  & 12.7\% \\
            & Triplet~\citep{wang2021solving}  & 27.5\% \\ \midrule
\multirow{4}{*}{Weakly Supervised} & W-Local CNN~\citep{wang2020weakly} & 28.8\% \\
            & W-MGN ~\citep{wang2020weakly}    & 29.5\% \\
            & WS-TAL~\citep{liu2023weakly}     & 34.4\% \\
            & CMIL (Ours)                             & 33.9\% \\ \bottomrule
\end{tabular}
\end{table}

The SYSU-30k dataset is very large 
and computationally expensive to optimize models on. 
Therefore, 
we compare directly to the results reported in prior work in Table~\ref{tab:sysu30k}. 
CMIL attains 33.9\% R1 accuracy outperforming the best transfer 
and self-supervised learning approaches by 10.3\% and 6.4\% respectively.
The best weakly supervised method is WS-TAL~\citep{liu2023weakly},
which is specifically engineered to optimize ReID models
when the labels are generated from video tracklets, 
matching the SYSU construction. 
WS-TAL reaches 34.4\% R1 accuracy. 
CMIL nearly matches this performance, 
lagging by only 0.5\%, 
using more general labeling assumptions.

Surprisingly, 
in every case, 
the alignment loss does not improve accuracy -- 
as shown in Table~\ref{tab:sweep-config}, 
$\gamma = 0$ and therefore the alignment loss is not used. 
During training, 
CMIL models are optimized at the bag level. 
Given a batch of bag representations, 
the model is optimized for bags with the same label to be close, 
and bags with a different label to be far from each other 
in representation space. 
The bag representations are built from crop representations, 
but nothing is preventing the bag and crop representations 
from being far apart. 
This should be problematic, 
because at test time, 
we are evaluating the quality of the crop embeddings. 

\begin{figure}[h]
    \centering
    \includegraphics[width=0.5\textwidth]{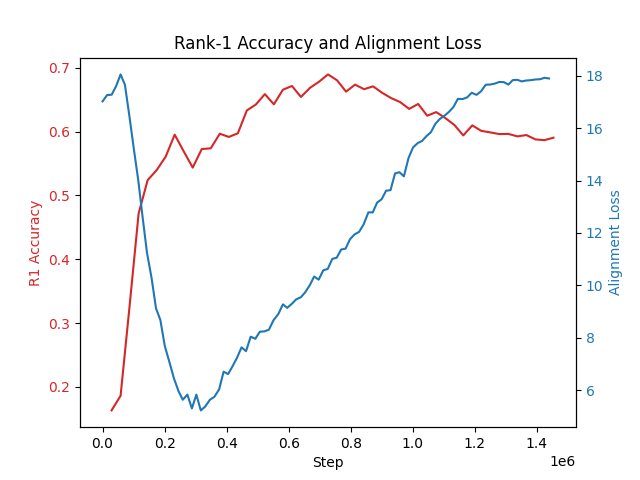}
    \caption{The rank-1 accuracy and the alignment loss throughout a training run. The alignment loss exhibits unintuitive behavior - the best alignment (i.e. lowest) does not correspond to the best model accuracy (i.e. highest). This behavior is characteristic of every model trained in this work, including those using different accumulation functions.}
    \label{fig:accuracy_vs_alignment}
\end{figure}

Figure~\ref{fig:accuracy_vs_alignment}
plots the rank-1 accuracy 
and alignment loss versus training step
when $\gamma=0$ (i.e. the alignment loss is not used). 
We see that
the bag and crop representations start quite different, 
and then begin growing more similar. 
However, 
there reaches a point relatively early in training
where the alignment between instance and bag representations begins to decrease. 
Interestingly, 
the performance of the model
(therefore the \emph{crop} embedding model) 
continues to improve, 
even while their embeddings diverge from bag embeddings.
This phenomenon is consistent across all of our experiments. 

It is not likely that 
this counterintuitive behavior is an artifact of 
the way we are measuring the alignment loss. 
It is not known which crop within a bag is
the crop representative of the bag label, 
so the alignment loss used here is the distance, 
using the same measure used during training, 
between the bag representation and the \emph{closest} crop
representation within the bag. 
A better understanding of this phenomenon requires further 
research. 

\begin{table*}[t]
\centering
\caption{Comparison of different accumulation functions on WL-Market-1501 and WL-MUDD datasets. Using a simple average of crop representations performs nearly as well as the set transformer.}
\label{tab:diff-acc-fns}
\begin{tabular}{@{}ccccccccc@{}}
    \toprule
                             & \multicolumn{4}{c}{WL-Market1501} & \multicolumn{4}{c}{WL-MUDD} \\ 
    Method                   & R1     & R5     & R10    & mAP    &  R1     & R5     & R10    & mAP   \\ \midrule
    CMIL w/Set Transformer   & 70.0\% & 86.4\% & 90.9\% & 48.8\% & 73.2\% & 90.0\% & 96.8\% & 75.1\%   \\
    CMIL w/Max               & 60.2\% & 78.7\% & 84.6\% & 33.9\% & 66.8\% & 82.1\% & 90.7\% & 68.2\% \\ 
    CMIL w/Avg               & 69.8\% & 85.6\% & 90.1\% & 44.1\% & 71.1\% & 88.6\% & 94.8\% & 74.5\% \\
    CMIL w/Sum               & 51.2\% & 72.0\% & 79.6\% & 24.3\% & 64.3\% & 79.3\% & 88.9\% & 62.4\% \\ \bottomrule
\end{tabular}
\end{table*}

\subsection{Ablation Study}
\label{sec:ablation}

In this ablation, 
we experiment with other, 
more traditional,
choices for permutation invariant accumulation function, 
specifically the max, average, and sum.
Each bag contains crops, 
and one or more of the crops in the bag are representative of the 
bag label. 
Of course, which crop specifically is unknown. 
Intuitively, 
the job of the accumulation function is to 
select the crop
(or a representation of the collection of crops)
that corresponds to the bag label. 

All aforementioned models have used a set transformer for the accumulation function, 
as the learnable attention-based model makes it possible to 
behave as a selector, 
or any arbitrary combination of the crop representations. 
However, it does come at the cost of complexity. 
Other reasonable, and much simpler, choices are 
to set the bag representation to be the max, average, or sum 
of the crop representations. 

In Table~\ref{tab:diff-acc-fns}, 
we compare the performance of the different accumulation functions
on both the WL-Market1501 and the WL-MUDD datasets. 
Interestingly, 
the average does well, 
matching, or nearly matching, the performance of the set transformer. 
This is surprising as the crops within a representative bag
of the bag label are the minority of samples, 
typically representing less than half of the samples within each bag. 
This could indicate that the set transformer is roughly just performing an average. 
A potential reason for this 
is that the non-representative crop features 
could act to cancel one another out
such that the bag representation is still close to the corresponding
representative features. 
Interestingly, 
tracking the alignment loss for each of these simpler 
accumulation functions shows the exact behavior as depicted in 
Figure~\ref{fig:accuracy_vs_alignment}.

\section{Related Work}

A large body of work has focused on supervised re-ID, 
where models are trained on data with individual object identity 
labels~\citep{ye2021deep, zheng2016person, zheng2017person, li2018harmonious, he2021transreid}. 
These approaches employ deep neural networks, 
to extract visual features
that are representative of specific identities, 
and optimize them according to
various loss functions,
including the identification loss where each identity is treated 
as a class~\citep{zheng2017person}, 
verification loss where pairwise relationships are optimized vi the contrastive loss~\citep{varior2016siamese}, 
triplet loss that treats the problem as 
a retrieval ranking problem~\citep{hermans2017defense}, 
and others have been proposed to optimize re-ID 
performance~\citep{chen2017beyond, yang2019patch, liu2019deep, song2019generalizable, wang2018mancs, chen2019self, zhou2019discriminative, xiao2017joint, guo2019beyond}. 
These methods perform well on baseline datasets, 
where ample data labeling is available. 

To alleviate the labeling bottleneck, 
recent works have begun investigating re-ID under weak supervision. 
Strategies include exploiting image-level labels~\citep{meng2019weakly}, 
pseudo-labels~\citep{wang2020weakly}, 
noisy label refinement~\citep{ye2021collaborative}, 
online captions~\citep{zhao2021weakly, guillaumin2010multiple}, 
and domain adaptation~\citep{yu2023weakly}. 
While showing promise, 
these methods still fall behind those 
that are fully supervised~\citep{zheng2021weakly}. 

Most similarly to our work, \citet{meng2019weakly}
leverage image-level labels in conjunction with 
Multiple Instance Learning (MIL) to effective facial recognition models.
MIL offers a paradigm to handle label ambiguity 
in training data by modeling labels at a bag level. 
A bag can be a collection of instances associated with a particular label, 
but we only know that one or more of the instances 
in that collection truly belongs to that label. 
Several works have adapted this specifically for treating video ``tracklets''
as a bag of instances~\citep{liu2023weakly, wang2021learning}
MIL has found diverse applications including 
image classification (particularly medical imagery)~\citep{wu2015deep, sudharshan2019multiple}, 
object detection~\citep{yuan2021multiple, wan2019c, huang2017salient},  
and drug discovery~\citep{fu2012implementation}. 
Critically, 
\citet{meng2019weakly} apply MIL to the person re-identification problem 
in the identity loss setting. 
In contrast, CMIL improves upon this by allowing for use of
contrastive learning, 
which has shown significant advantages in person ReID 
and related settings~\citep{hermans2017defense, garg2023complementary}.

Lastly, 
we must mention the work in unsupervised ReID \citep{Fu_2021_CVPR, Li_2018_ECCV, Wang_2020_CVPR, lin2019bottom, Yu_2019_CVPR, fan2018unsupervised}. 
These methods do not require labels. 
Typically, 
these methods use iterative clustering and classification, such that unlabeled images are clustered into ``pseudo'' classes, 
which are then used to train or update a model. 
Then the new/updated model is used to refine the pseudo labels, 
and so on. 
Improvements to this standard approach include 
substituting the clustering step for pairwise comparisions~\citep{lin2020unsupervised}, 
and an improved clustering step by improving the global clusters 
using ensembles of image-part based predictions~\citep{cho2022part}. 
Of course, 
performance is still greatly improved 
when labels are present~\citep{xiang2023rethinking, zhu2022pass, luo2021self, yang2022unleashing, Fu_2022_CVPR}.

\section{Conclusion}

In this paper, 
we introduced Contrastive Multiple Instance Learning (CMIL), 
a novel framework tailored for more effective person re-identification 
under weak supervision. 
CMIL tackles the challenge of learning discriminative person representations 
when only bag-level labels indicating a shared identity 
among a group of photos are available. 
Although the model is trained at the bag level, 
the person-level representations 
improve alongside the quality of the bag-level representations. 
We experiment with adding an alignment loss 
to further encourage the person and bag representations to be similar, 
but found it ineffective empirically. 

We experiment on three datasets, 
one of which is the Weakly Labeled Muddy Racer Re-Identification Dataset 
(WL-MUDD), which is curated and released from 
real-world weak labels from PerformancePhoto.co. 
Across these experiments, 
CMIL consistently achieved state-of-the-art rank-1, 
rank-5, and rank-10 accuracy as well as mean average precision. 
On the large-scale SYSU-30k dataset, 
CMIL matched the top-reported result 
while requiring fewer assumptions. 
Ablations also revealed surprising effectiveness 
of average pooling for instance aggregation, 
suffering only slight performance degradation to the set transformer.

The contributions of this work are threefold. 
First, 
we introduce the new Weakly Labeled Muddy Racer Re-Identification dataset 
(WL-MUDD) built from \url{PerformancePhoto.co}, 
an off-road photograph platform. 
Second, 
we introduce the CMIL framework
that enables efficient exploitation of cheap weak supervision 
for person re-id through enabling contrastive learning 
with Multiple Instance Learning. 
And third, 
we show the efficacy of CMIL on two real-world datasets
and one synthetic, outperforming baselines.

\bibliographystyle{ACM-Reference-Format}
\bibliography{references}

\end{document}